\documentclass{article}

\usepackage[final]{neurips_2021}

\usepackage[utf8]{inputenc} % allow utf-8 input
\usepackage[T1]{fontenc}    % use 8-bit T1 fonts
\usepackage{hyperref}       % hyperlinks
\usepackage{url}            % simple URL typesetting
\usepackage{booktabs}       % professional-quality tables
\usepackage{amsfonts}       % blackboard math symbols
\usepackage{nicefrac}       % compact symbols for 1/2, etc.
\usepackage{microtype}      % microtypography
\usepackage{xcolor}         % colors
\setcitestyle{numbers}
\setcitestyle{square}
\setcitestyle{comma}
\usepackage{bm}
\usepackage{graphicx}       % for putting images
\usepackage{wrapfig}
\usepackage{lipsum}
\usepackage[hang,flushmargin]{footmisc}
\usepackage{amsmath}
\newcommand{\boldsym}[1]{\bm{#1}}
\newcommand{\scene}[1]{\textit{#1}}
\newcommand{\Tau}{\mathcal{T}}
\newcommand{\diff}{\mathop{}\!\mathrm{d}}

\title{Neural Relightable Participating Media Rendering}

\author{
  Quan Zheng$^{1,2}$, Gurprit Singh$^{1}$, Hans-Peter Seidel$^{1}$   \\
  $^{1}$Max Planck Institute for Informatics, 66123 Saarbr{\"u}cken, Germany\\
  $^{2}$Institute of Software, Chinese Academy of Sciences, 100190 Beijing, China\\
  \texttt{\{qzheng, gsingh, hpseidel\}@mpi-inf.mpg.de} \\
}

\begin{document}

\maketitle

\begin{abstract}
  Learning neural radiance fields of a scene has recently allowed realistic novel view synthesis of the scene, but they are limited to synthesize images under the original fixed lighting condition. Therefore, they are not flexible for the eagerly desired tasks like relighting, scene editing and scene composition. To tackle this problem, several recent methods propose to disentangle reflectance and illumination from the radiance field. These methods can cope with solid objects with opaque surfaces but participating media are neglected. Also, they take into account only direct illumination or at most one-bounce indirect illumination, thus suffer from energy loss due to ignoring the high-order indirect illumination. We propose to learn neural representations for participating media with a complete simulation of global illumination. We estimate direct illumination via ray tracing and compute indirect illumination with spherical harmonics. Our approach avoids computing the lengthy indirect bounces and does not suffer from energy loss. Our experiments on multiple scenes show that our approach achieves superior visual quality and numerical performance compared to state-of-the-art methods, and it can generalize to deal with solid objects with opaque surfaces as well.
\end{abstract}

\section{Introduction}

From natural phenomenons like fog and cloud to ornaments like jade artworks and wax figures, participating media objects are pervasive in both real life and virtual content like movies or games. Inferring the bounding geometry and scattering properties of participating media objects from observed images is a long-standing problem in both computer vision and graphics. Traditional methods addressed the problem by exploiting specialized structured lighting patterns~\cite{gu2008compressive, hawkins2005acquisition, fuchs2007density} or using discrete representations~\cite{hasinoff2007photo}. These methods, however, require the bounding geometry of participating media objects to be known.

Learning neural radiance fields or neural scene representations~\cite{mildenhall2020nerf, lombardi2019neural, sitzmann2019scene} has achieved remarkable progress in image synthesis. They are able to optimize the representations with the assistance of a differentiable ray marching process. However, these methods are mostly designed for novel view synthesis and have baked in materials and lighting into the radiance fields or surface color. Therefore, they can hardly support downstream tasks such as relighting and scene editing. Recent work~\cite{bi2020neural, srinivasan2020nerv} has taken initial steps to disentangle the lighting and materials from radiance. For material, their methods are primarily designed for solid objects with opaque surfaces, thus they assume an underlying surface at each point with a normal and a BRDF. The assumed prior, however, does not apply to non-opaque participating media which has no internal surfaces. For lighting, neural reflectance field~\cite{bi2020neural} simulates direct illumination from a single point light, whereas NeRV~\cite{srinivasan2020nerv} handles direct illumination and one-bounce indirect illumination. They generally suffer from the energy loss issue due to ignoring the high-order indirect illumination. However, indirect lighting from multiple scattering plays a significant role in the final appearance~\cite{mukaigawa2010analysis} of participating media. 

In this paper, we propose a novel neural representation for learning relightable participating media. Our method takes as input a set of posed images with varying but known lighting conditions and designs neural networks to learn a disentangled representation for the participating media with physical properties, including volume density, scattering albedo and phase function parameter. To synthesize images, we embed a differentiable physically-based ray marching process in the framework. In addition, we propose to simulate global illumination by embedding the single scattering and multiple scattering estimation into the ray marching process, where single scattering is simulated by Monte Carlo ray tracing and the incident radiance from multiple scattering is approximated by spherical harmonics (SH). Without supervising with ground-truth lighting decomposition, our method is able to learn a decomposition of direct lighting and indirect lighting in an unsupervised manner. 

Our comprehensive experiments demonstrate that our method achieves better visual quality and higher numerical performance compared to state-of-the-art methods. Meanwhile, our method can generalize to handle solid objects with opaque surfaces. We also demonstrate that our learned neural representations of participating media allow relighting, scene editing and insertion into another virtual environment. To summarize, our approach has the following contributions:
\begin{enumerate}
\item {We propose a novel method to learn a disentangled neural representation for participating media from posed images and it is able to generalize to solid objects.}
\item {Our method deals with both single scattering and multiple scattering and enables the unsupervised decomposition of direct illumination and indirect illumination.}
\item {We demonstrate flexibility of the learned representation of participating media for relighting, scene editing and scene compositions.}
\end{enumerate}

\section{Related Work}

\paragraph{Neural scene representations.} 
Neural scene representations~\cite{eslami2018neural, sitzmann2019scene, lombardi2019neural} are important building blocks for the recent progress in synthesizing realistic images. Different from representations of components such as ambient lighting and cameras~\cite{satilmis2016machine, zheng2017neur, zheng2017adaptive} of scenes, neural scene representation~\cite{eslami2018neural} learns an embedding manifold from 2D images and 
scene representation networks~\cite{sitzmann2019scene} aim to infer the 3D context of scenes from images. Classic explicit 3D representations, such as voxels~\cite{sitzmann2019deepvoxels, peng2020convolutional, lombardi2019neural}, multiplane images~\cite{flynn2019deepview, zhou2018stereo} and proxy geometry~\cite{zhang2021neural} are exploited to learn neural representations for specific purposes. These explicit representations generally suffer from the intrinsic resolution limitation. To sidestep the limitation, most recent approaches shift towards implicit representations, like signed distance fields~\cite{park2019deepsdf, chabra2020deep}, volumetric occupancy fields~\cite{chen2019learning, mescheder2019occupancy, liu2020neural, saito2019pifu}, or coordinate-based neural networks~\cite{xu2019disn, sitzmann2019scene, sitzmann2020implicit, zheng2020neural}. By embedding a differentiable rendering process like ray marching~\cite{lombardi2019neural, mildenhall2020nerf} or sphere tracing~\cite{sitzmann2019scene, liu2020dist} into these implicit representations, these methods are capable of optimizing the scene representations from observed images and synthesizing novel views after training. While they generally show improved quality compared to interpolation based novel view synthesis methods~\cite{srinivasan2017learning, srinivasan2019pushing}, the learned representations are usually texture colors and radiance, without separating lighting and materials. By contrast, we propose to learn a neural representation with disentangled volume density, scattering properties and lighting, which allow the usages in relighting, editing and scene composition tasks.

\paragraph{Volume geometry and properties capture.} Acquiring geometry and scattering properties of participating media has long been of the interest to the computer vision and graphics community. Early methods utilize sophisticated scanning and recording devices~\cite{goesele2004disco, hawkins2005acquisition} and specialized lighting patterns~\cite{fuchs2007density, gu2008compressive} to capture volume density. Computational imaging methods~\cite{gkioulekas2016evaluation, gkioulekas2013inverse} frame the inference of scattering properties from images as an inverse problem, but they require that the geometries of objects are known. Based on the differentiable path tracing formulation~\cite{azinovic2019inverse}, the inverse transport method~\cite{che2020towards} incorporates a differentiable light transport~\cite{zhang2019differential} module within an analysis-by-synthesis pipeline to infer scattering properties, but it aims for known geometries and homogeneous participating media. In contrast, our method learns the geometries and scattering properties of participating media simultaneously and our method can deal with both homogeneous and heterogeneous participating media.

\paragraph{Relighting.} Neural Radiance Field (NeRF)~\cite{mildenhall2020nerf} and its later extensions~\cite{dellaert2020neural} encode the geometry and radiance into MLPs and leverage ray marching to synthesize new views. While they achieve realistic results, they are limited to synthesize views under the same lighting conditions as in training. To mitigate this, appearance latent code~\cite{martin2020nerf, schwarz2020graf} are used to condition on the view synthesis. Recent approaches~\cite{bi2020neural, srinivasan2020nerv} decompose materials and lighting by assuming an opaque surface at each point, but this does not apply to participating media. After training, density and materials can be extracted~\cite{boss2020nerd} to render new views using Monte Carlo methods~\cite{jarosz2011comprehensive, zheng2015visual, novak2018monte}.
Instead, our method models participating media as a field of particles that scatter and absorb light, which are in accordance with its nature. Neural reflectance field~\cite{bi2020deep, bi2020neural} requires collocated cameras and lights during training and simulates only direct illumination. NeRV~\cite{srinivasan2020nerv} and OSF~\cite{guo2020object} simulate merely one-bounce indirect light transport because of the prohibitive computation cost for long paths. However, ignoring the high-order indirect illumination leads to the potential problem of energy loss. By contrast, we use Monte Carlo ray tracing to compute direct lighting and propose to learn a spherical harmonic field for estimating the complete indirect lighting. The PlenOctree~\cite{yu2021plenoctrees} uses spherical harmonics to represent the outgoing radiance field as in NeRF, but it does not allow relighting. With both direct illumination and indirect illumination properly estimated, our method enables a principled disentanglement of volume density, scattering properties, and lighting.
\section{Background} \label{sec:background}

\paragraph{Volume rendering.} The radiance carried by a ray after its interaction with participating media can be computed based on the radiative transfer equation~\cite{lafortune1996rendering}
\begin{equation} \label{eq:radiance}
L_{o}\left(\boldsym{r}_{0}, \boldsym{r}_{d}\right) = \int_{0}^{\infty}\tau\left(\boldsym{r}(t)\right)\sigma\left(\boldsym{r}(t)\right)L\left(\boldsym{r}(t), -\boldsym{r}_{d}\right)\diff t.
\end{equation}
Here, $\boldsym{r}$ is a ray starting from $\boldsym{r}_{0}$ along the direction $\boldsym{r}_{d}$ and $\boldsym{r}(t)=\boldsym{r}_{o}+t\cdot{r}_{d}$ denotes a point along the ray at the parametric distance\footnote{While the integration accounts for a $t$ going to $\infty$, $t$ covers only the range with participating media in practice.} $t$. $L_{o}$ is the received radiance at $\boldsym{r}_{0}$ along $\boldsym{r}_{d}$. $\sigma$ denotes the extinction coefficient that is referred to as volume density. The $\tau\left(\boldsym{r}(t)\right)$ is the transmittance between $\boldsym{r}(t)$ and $\boldsym{r}_{0}$ and it can be computed by $\exp\left(-\int_{0}^{t}\sigma(\boldsym{r}(s)) \diff s\right)$. The $L(\cdot)$ inside the integral stands for the in-scattered radiance towards $\boldsym{r}_{0}$ along $-\boldsym{r}_{d}$. NeRF~\cite{mildenhall2020nerf} models the in-scattered radiance $L$ (Eq.~\ref{eq:radiance}) as a view dependent color $c$, but it ignores the underlying scattering event and incident illumination. Since the learned radiance field of NeRF bakes in the lighting and materials, it allows merely view synthesis under the original fixed lighting, without the support for relighting.

\paragraph{Ray marching.} The integral in Equation~\ref{eq:radiance} can be solved with the numerical integration method, \textit{ray marching}~\cite{kniss2003model}. This is generally done by casting rays into the volume and taking point samples along each ray to collect volume density and color values~\cite{mildenhall2020nerf, lombardi2019neural}. The predicted color of a ray is computed by $L_{o}\left(\boldsym{r}_{0}, \boldsym{r}_{d}\right) = \sum_{j}\tau_{j}\cdot\alpha_{j}\cdot L\left(\boldsym{r}(t_{j}), -\boldsym{r}_{d}\right)$, where $\alpha_j = 1-\exp\left(-\sigma_{j}\cdot\delta_{j}\right)$, $\delta_{j} = \Vert t_{j+1} - t_{j}\Vert_{2}$ and $\tau_{j}=\prod_{i=1}^{j-1}(1-\alpha_{i})$.

\section{Neural Relightable Participating Media}

In this work, we aim to learn neural representations of participating media with disentangled volume density and scattering properties. We model the participating media as a field of particles that absorb and scatter light-carrying rays. Below we first describe our disentangled neural representation based on a decomposition with single scattering and multiple scattering. Then, we depict our neural network design, followed by details of a volume rendering process for synthesizing images and the details of training.

\subsection{Lighting Decomposition} \label{sec:lightdecompose}
We firstly write the in-scattered radiance $L\left(\boldsym{r}(t), -\boldsym{r}_{d}\right)$ in Equation~\ref{eq:radiance} as an integral of light-carrying spherical directions over a $4\pi$ steradian range
\begin{equation} \label{eq:expandL}
    L\left(\boldsym{r}(t), -\boldsym{r}_{d}\right) = \int_{\Omega_{4\pi}}S\left(\boldsym{r}(t), -\boldsym{r}_{d}, \omega_{i}\right)L_{in}(\boldsym{r}(t), \omega_{i})\diff\omega_{i},
\end{equation}
where $L_{in}$ is the incident radiance at $\boldsym{r}(t)$ from the direction $\omega_{i}$. $S(\cdot)$ is a scattering function which determines the portion of lighting that is deflected towards the $-\boldsym{r}_{d}$. Previous methods~\cite{bi2020neural,srinivasan2020nerv, boss2020nerd} assume a surface prior with a normal at every point and account for $2\pi$ hemispherical incident directions. Accordingly, they define the scattering function $S$ as a BRDF function.
This assumption, however, can hardly match the participating media objects which have no internal surfaces and normals. By contrast, we deal with light-carrying directions over the full $4\pi$ steradian range.
Specifically, we define $S = a(\boldsym{r}(t))\cdot\rho(-\boldsym{r}_{d}, \omega_{i}, g)$ to account for the scattering over the full spherical directions. Here, $a(\boldsym{r}(t))$ is the scattering albedo. $\rho$ is the Henyey-Greenstein (HG)~\cite{henyey1941diffuse} phase function\footnote{Both the incident and outgoing directions point away from a scattering location in this paper.} that decides the scattering directionality (Appendix A), 
where $g$ is an \textit{asymmetry parameter} in $(-1, 1)$. For brevity, we omit $g$ in the notation of $\rho$. Then, we can rewrite the radiance of ray $\boldsym{r}$ as the disentangled form:
\begin{equation} \label{eq:inscatter}
L_{o}\left(\boldsym{r}_{0}, \boldsym{r}_{d}\right) = \int_{0}^{\infty}\tau\left(\boldsym{r}(t)\right)\sigma\left(\boldsym{r}(t)\right)\int_{\Omega_{4\pi}}a(\boldsym{r}(t))\rho(-\boldsym{r}_{d}, \omega_{i})L_{in}(\boldsym{r}(t), \omega_{i})\diff\omega_{i}\diff t,
\end{equation}
from which the volume density $\sigma$ decides the geometries, and the albedo $a$ along with the phase function $\rho$ controls the scattering of rays. We propose to train neural networks to learn the volume density and scattering properties.

To compute $L_{o}$, we additionally need to estimate $L_{in}$ (Eq.~\ref{eq:inscatter}). This can be conducted by recursively substituting the $L_{o}$ into $L_{in}$ and expressing the radiance for a pixel $k$ as an integral over all paths $P_{k} = \int_{\Psi}g_{k}(\bar{x})d\mu(\bar{x})$, where $\bar{x}$ is a light-carrying path, $g_{k}$ denotes a measurement contribution function, $\mu$ is the measure for the path space, and $\Psi$ is the space of paths with all possible lengths. $P_{k}$ is then computed by a summation of the contributions of all paths. The contribution of a path with length $i$ can be written as
\begin{equation} \label{eq:path}
P_{k,i}=\int_{l_{1}}\int_{\Omega}\cdots\int_{l_{i-1}}\int_{\Omega} L_{e}\left(x(t_{i-1})\right)V\left(x(t_{i-1}),\omega_{i-1}\right)\Tau(P_{k,i})\diff t_{1}\diff\omega_{1}\cdots\diff t_{i-1}\diff \omega_{i-1},
\end{equation}
where $x(t_{i-1})$ is a point along the {\small{$(i-1)$}}-th ray segment with the length $l_{i-1}$, $\omega_{i-1}$ is the scattering direction in the space $\Omega$ and $\omega_{0} = \boldsym{r}_{d}$, $L_{e}$ denotes the emitted radiance towards $x(t_{i-1})$ from a light source, $V$ accounts for the transmittance between $x(t_{i-1})$ and the light source, $\Tau(P_{k, i}) = \prod_{j=1}^{i-1}V(x(t_{j}), \omega_{j}) \sigma(x(t_{j}))\cdot\prod_{j=1}^{i-1}\rho(-\omega_{j-1}, \omega_{j})a(x(t_{j}))$ is called the path throughput. Volumetric path tracing~\cite{novak2018monte} utilizes Monte Carlo method to estimate the integral but its computational cost increases quickly when using high sampling rates of paths with many bounces. To reduce the cost, NeRV~\cite{srinivasan2020nerv} truncates the paths and only considers up to one indirect bounce, namely $i=3$ in Eq.~\ref{eq:path}. This, however, leads to energy loss since high-order indirect illumination are neglected. 

Instead of tracing infinitely long paths or truncating the paths, we propose to decompose the in-scattered radiance $L$ in Eq.~\ref{eq:radiance} as $L = L_s + L_m$, where $L_s$ is the single-scattering contribution and $L_m$ denotes the multiple-scattering contribution. Therefore, $L_{o}$ can be split into two integrals: $L_{o, s}=\int_{0}^{\infty}\tau\left(\boldsym{r}(t)\right)\sigma\left(\boldsym{r}(t)\right)L_s\diff t$ and $L_{o, m}=\int_{0}^{\infty}\tau\left(\boldsym{r}(t)\right)\sigma\left(\boldsym{r}(t)\right)L_m\diff t$, that can be evaluated separately. 

\paragraph{Single scattering.} To compute the $L_{o, s}$, we evaluate the following integral at each sample point $\boldsym{r}(t)$ along the camera ray $\boldsym{r}$
\begin{equation}\label{eq:singlescatter}
L_s =\int_{\Omega_{4\pi}}a(\boldsym{r}(t))\rho(-\boldsym{r}_{d}, \omega_{i})L_{e}(\boldsym{r}(t), \omega_{i})V(\boldsym{r}(t), \omega_{i}) \diff\omega_{i}.
\end{equation}
Here, $L_e$ is the emitted radiance from a light source to the point $\boldsym{r}(t)$ and $V$ is the transmittance between $\boldsym{r}(t)$ and a light source. The transmittance can be computed with another integral of volume density as described in Section~\ref{sec:background}, but computing the integral for all points leads to high computation cost during training and inference. Therefore, we train a visibility neural network to regress the transmittance value as done in~\cite{srinivasan2020nerv}.

\paragraph{Multiple scattering.} For $L_{o,m}$, we evaluate the $L_m=\int_{\Omega_{4\pi}}a(\boldsym{r}(t))\rho(-\boldsym{r}_{d}, \omega_{i})L_{in}(\boldsym{r}(t), \omega_{i})d\omega_{i}$. $L_m$ aggregates the incoming radiance of rays that have been scattered at least once in the participating media. Since the distribution of incident radiance from multiple scattering is generally smooth, we propose to represent the incident radiance $L_{in}$ as a spherical harmonics expansion: $L_{in}(\omega_i) = \mathcal{F}\left(\sum_{l=0}^{l_{max}}\sum_{m=-l}^{l}c_{l}^{m}Y_{l}^{m}(\omega_i)\right)$, where $l_{max}$ is the maximum spherical harmonic \textit{band}, $c_{l}^{m}\in\mathbb{R}^3$ are spherical harmonic coefficients for the RGB spectrum, $Y_{l}^{m}$ are spherical harmonic basis functions and $\mathcal{F}(x)=\max(0, x)$. Therefore, we compute the multiple-scattering contribution with
\begin{equation}\label{eq:multiplescatter}
L_{m}=\int_{\Omega_{4\pi}}a(\boldsym{r}(t))\rho(-\boldsym{r}_{d}, \omega_{i})\cdot\mathcal{F}\left(\sum_{l=0}^{l_{max}}\sum_{m=-l}^{l}c_{l}^{m}Y_{l}^{m}(\omega_i)\right)d\omega_{i}.
\end{equation}
\begin{wrapfigure}{R}{0.51\textwidth}
\centering
\vspace{-1em}
\includegraphics[width=1.0\linewidth]{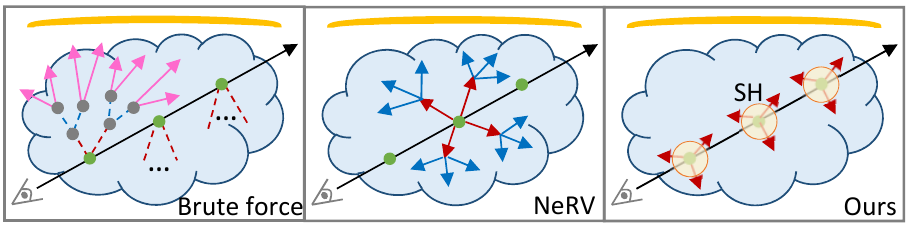}
\vspace{-1em}
\caption{\label{fig:rayexplosion}Visualization of path bounces.}
\vspace{-1em}
\end{wrapfigure}
We employ a neural network to learn spherical harmonic coefficients. By using spherical harmonics for the incident radiance from multiple scattering, we sidestep the lengthy extension of the path integral (Eq.~\ref{eq:path}). 
Since we introduce the approximation of multiple scattering at the primary rays, we sidestep the explosion of rays.
Figure~\ref{fig:rayexplosion} visualizes the explosion of rays when computing indirect illumination under an environment lighting. Multiple shadow rays are needed to account for the directional emission from the light source. The brute-force ray splitting approach leads to explosion of rays and is impractical. NeRV reduces the ray count by tracing up to one indirect bounce, but it has a complexity of $\mathcal{O}(M\cdot N)$, where $M$ is the number of first indirect bounces (red) and $N$ is the number of shadow rays (blue). Our method uses spherical harmonics to handle indirect illumination as a whole and its complexity is $\mathcal{O}(k\cdot M)$, where $k$ is the sampling rates along the primary rays.
\begin{figure*}[t]
\centering
\includegraphics[width=0.95\linewidth]{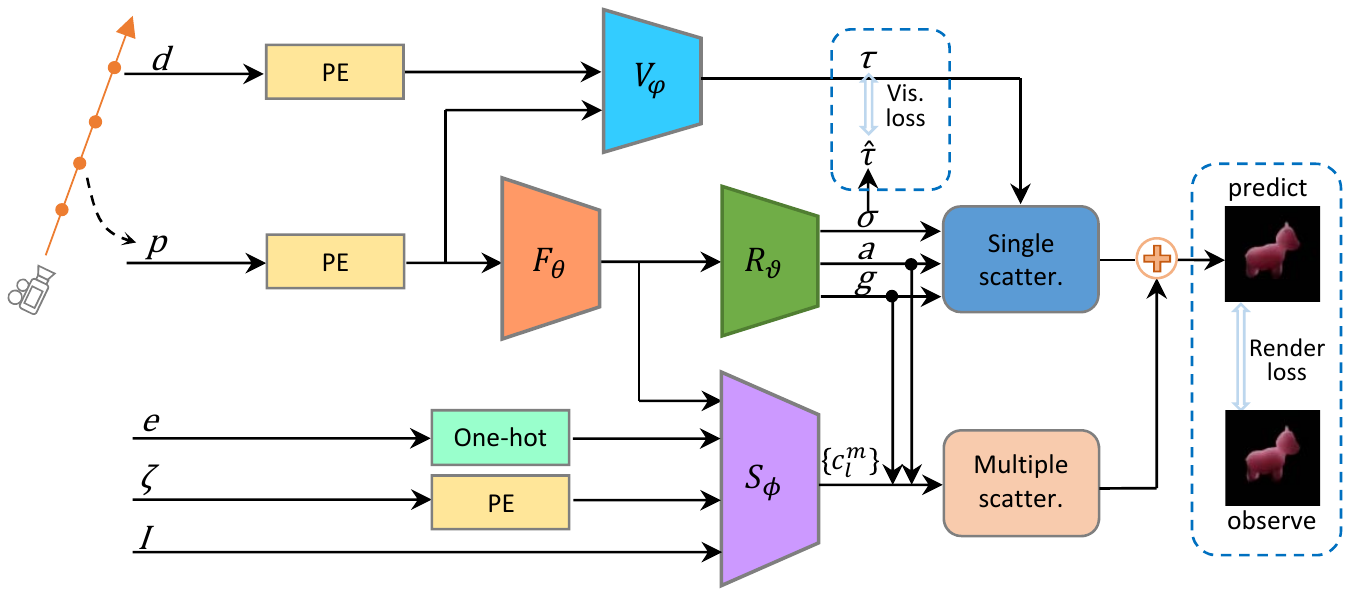}
\caption{\label{fig:overview} Our overall architecture for learning neural participating media. ``PE'' denotes the positional encoding and ``one-hot'' denotes the one-hot encoding. The rendering loss and the visibility loss correspond to the summands in Eq.~\ref{eq:total_loss}.
}
\end{figure*}
\paragraph{Network architectures.}
Figure~\ref{fig:overview} presents our overall architecture. Our neural networks are based on the coordinate-based MLP strategy, and we use frequency-based positional encoding~\cite{ mildenhall2020nerf,tancik2020fourier} $E$ to map an input coordinate $p$ to a higher dimensional vector $E(p)$ before sending it to the neural networks. Specifically, we use the MLP $R_{\vartheta}$ to predict volume density $\sigma$ (1D), scattering albedo $a$ (3D), and \textit{asymmetry parameter} $g$ (1D). Meanwhile, we employ the MLP $S_{\phi}$ to learn spherical harmonic coefficients $c_{l}^{m}$. Here, $S_{\phi}$ is conditioned on the point light location $\zeta$, the point light intensity $I$, and a binary indicator $e$ which is one-hot encoded to indicate the existence of environment lighting.
Since both the property network $R_{\vartheta}$ and the SH network $S_{\phi}$ takes as input the encoded coordinate, we introduce a feature network $F_{\theta}$ to predict shared features for downstream MLPs. To get the visibility for shadow rays, we train another MLP $V_{\varphi}$, which takes in the encoded direction $d$ in addition to the encoded coordinate $p$, to learn visibility values for the estimation of single scattering. In summary, we have
\begin{equation}
R_{\vartheta}: F_{\theta}(E(p))\rightarrow (\sigma, a, g)\quad
S_{\phi}: (e, \zeta, I, F_{\theta}(E(p)))\rightarrow \{c_{l}^{m}\}\quad
V_{\varphi}: (\boldsym{r}(t), E(d))\rightarrow \tau.
\end{equation}
Note that the above $R_{\vartheta}$ learns per-location \textit{asymmetry parameter} $g$ (Appendix A). Yet, for scenes with a single participating media object, we use a singe $g$ and optimize it during training.

\subsection{Volume Rendering}\label{sec:rendering}
Based on the above decomposition, we employ \textit{ray marching} (Sec.~\ref{sec:background}) to numerically estimate $L_{o, s}$ and $L_{o, m}$. Hence, the final radiance $L_o$ of the camera ray $\boldsym{r}$ in Eq.~\ref{eq:radiance} can be computed as:
\begin{equation}
    L_{o}\left(\boldsym{r}\right) =\Sigma_{j=1}^{N}\tau\left(\boldsym{r}(t)\right)\left(1-\exp\left(-\sigma\left(\boldsym{r}(t)\right)\cdot\delta t_{j}\right)\right)\left(L_s + L_m\right),
\end{equation}
where we sample $N=64$ query points in a stratified way along the ray $\boldsym{r}$ and $\delta t_{j}=\Vert\boldsym{r}(t_{j+1})-\boldsym{r}(t_{j})\Vert_{2}$ is the step size. For each point sample, we query the MLPs to obtain its scattering properties and SH coefficients for computing single scattering and multiple scattering.

We compute single scattering at a point based on Eq.~\ref{eq:singlescatter}. We shoot shadow rays towards light sources to get the emitted radiance $L_{e}$ according to the light types. For environment lighting, we sample $64$ directions stratified over a sphere around the point to obtain incident radiance. For a point light, we directly connect the query point to the light source. To account for the attenuation of light radiance, we query $V_{\varphi}$ to get the visibility to the light source from the query point.

For multiple scattering,  we uniformly sample $K=64$ random incident directions over the sphere around each query point, evaluate the incident radiance $L_{in}$ along each direction using the learned spherical harmonic coefficients, and estimate the integral in  Eq.~\ref{eq:multiplescatter} with a Monte Carlo integration $L_m=1/K\sum_{i=1}^{K}a\rho(\omega_{i})L_{in}(\omega_i)$. For brevity, we omit the $\boldsym{r}$ notation. 
Note that the visibility towards the light source is not needed in the computation.

\subsection{End-to-end Learning}
Based on the fully differentiable pipeline, we can end-to-end learn a neural representation for each scene. The learning requires a set of posed RGB images and their lighting conditions. During each training iteration, we trace primary rays through the media. Along each primary ray, we estimate single scattering using shadow rays and compute multiple scattering contribution via the learned spherical harmonic coefficients as described in Sec.~\ref{sec:rendering}. We optimize the parameters of $F_{\theta}$, $R_{\vartheta}$, and $S_{\phi}$ by minimizing a rendering loss between the predicted radiance $L_{o}(\boldsym{r})$ from \textit{ray marching} and the radiance $\hat{L}_{o}(\boldsym{r})$ from input images. To train the visibility network $V_{\varphi}$, we use the transmittance $\hat{V}_{\vartheta}$ computed from the learned volume density as the ground truth and minimize the visibility loss between the prediction $V_{\varphi}$ and the ground truth. Therefore, our loss function includes two parts:
\begin{equation}\label{eq:total_loss}
\mathcal{L} = \sum_{\boldsym{r}\in\mathcal{R}}\Vert\Gamma({L}_{o}(\boldsym{r}))-\Gamma(\hat{L}_{o}(\boldsym{r}))\Vert_{2}^{2} + \mu\cdot\sum_{\boldsym{r}\in\mathcal{R}, t}\Vert V_{\varphi}(\boldsym{r}(t), \boldsym{r}_{d})-\hat{V}_{\vartheta}(\boldsym{r}(t), \boldsym{r}_{d})\Vert_{2}^{2},
\end{equation}
where $\Gamma(L) = L/(1+L)$ is a tone mapping function, $\mathcal{R}$ is a batch of camera rays and $\mu=0.1$ is the hyperparameter to weight the visibility loss.

\subsection{Implementation Details}\label{sec:implementation}
Our feature MLP $F_{\theta}$ has 8 fully-connected ReLU (FC-ReLU) layers with 256 channels per layer. The downstream $S_{\phi}$ consists of 8 FC-ReLU layers with 128 channels per layer, whereas the $R_{\vartheta}$ uses one such layer with 128 channels. The visibility MLP $V_{\varphi}$ has 4 FC-ReLU layers with 256 channels per layer. We set the maximum positional encoding frequency to $2^{8}$ for coordinates $p$, $2^{1}$ for directions $d$, and $2^{2}$ for the 3D location of the point light. 

We train all neural networks together to minimize the loss (Eq.~\ref{eq:total_loss}). We use the Adam~\cite{kingma2014adam} optimizer with its default hyperparameters and schedule the learning rate to decay from $1\times 10^{-4}$ to $1\times 10^{-5}$ over 200K iterations. For each iteration, we trace a batch of $1200$ primary rays. Note we stop the gradients from the visibility loss to the property network and the feature network so that they do not compromise the learning to match the visibility network.
\section{Experiments}

We firstly evaluate our method by comparing it with state-of-the-art methods on simultaneous relighting and view synthesis. Then, we demonstrate that our learned neural representations allow flexible editing and scene compositions, followed by ablation studies of this approach. Please refer to the appendices for additional results.

\subsection{Experiment Settings}\label{sec:exp_setting}

\paragraph{Compared methods.} We compare our method with state-of-the-art baselines~\cite{bi2020neural, srinivasan2020nerv}. They are designed for scenes with solid objects and do not trivially extend to handle participating media, so we implement them with the new functionality to handle participating media. Please refer to the Appendix D for the implementation details. 
\paragraph{Datasets.} We produce datasets from seven synthetic participating media scenes. The \scene{Cloud} scene is heterogeneous media and the others are homogeneous media. \scene{Bunny4-VaryA} and \scene{Buddha3} are set with spatially varying albedo. \scene{Bunny4-VaryG} and \scene{Buddha3} are configured with spatially varying asymmetry parameters.
Each scene is individually illuminated with two lighting conditions. The first one ``point'' has a white point light with varying intensities sampled within $50\sim 900$ and its location is randomly sampled on spheres with the radius ranging from $3.0$ to $5.0$; The second one ``env + point'' contains a fixed environment lighting and a randomly sampled white point light. Each dataset contains $180$ images, from which we use $170$ images for training and the remaining for validation. In addition, we prepare a test set with $30$ images for each scene to test the trained models. Each test image is rendered with a new camera pose and a new white point light that is located on a sphere of the radius $4.0$. Since Bi's method~\cite{bi2020neural} requires a collocated camera and light during training, we additionally generated such datasets for it. For the ``env + point'' datasets, we randomize the usage of environment lighting across images and record a binary indicator for each image.

\subsection{Results}\label{sec:results}
\begin{figure}[t]
\centering
\includegraphics[width=1.0\textwidth]{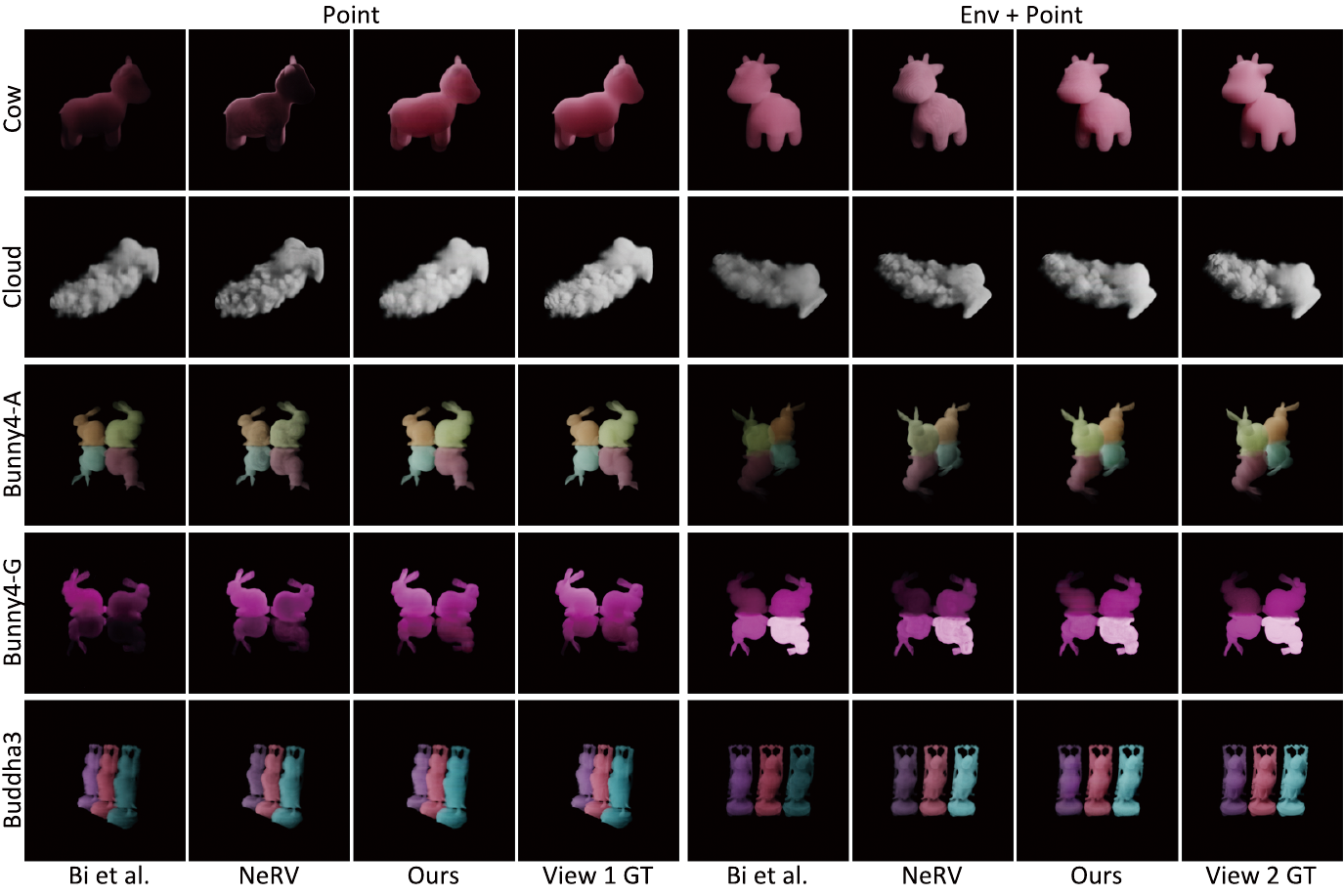}
\vspace{-1.5em}
\caption{\label{fig:relight_compare} Qualitative comparisons of simultaneous view synthesis and relighting. The training illumination for the left half is ``point'' which contains a single point light. The training illumination for the right half is ``env + point''. GT denotes the ground truth image.}
\end{figure}

\begin{table}
\begin{center}
\small
\caption{\label{tab:point_quality} Quantitative comparisons on the test data for training on the ``point'' illumination. We measure image qualities with PSNR ($\uparrow$), SSIM ($\uparrow$) and ELPIPS ($\downarrow$)~\cite{kettunen2019lpips}. ELPIPS values below have a scale of $\times 10^{-2}$. Note the tabulated values are the mean values over all images of a test set. }
\vspace{-0.5em}
\setlength{\tabcolsep}{3pt}
\resizebox{\textwidth}{!}{
\begin{tabular}{l ccc ccc ccc ccc ccc} % lccc ccc ccc ccc ccc
\toprule
{\small{Point}}& \multicolumn{3}{c}{Cow}  &  \multicolumn{3}{c}{Cloud} & \multicolumn{3}{c}{Bunny4-VaryA} & \multicolumn{3}{c}{Bunny4-VaryG} & \multicolumn{3}{c}{Buddha3}\\
\cmidrule(lr){2-4}\cmidrule(lr){5-7}\cmidrule(lr){8-10}\cmidrule(lr){11-13}\cmidrule(lr){14-16}
Method & PSNR & SSIM &ELPIPS & PSNR  & SSIM &ELPIPS &PSNR & SSIM & ELPIPS &PSNR & SSIM & ELPIPS&PSNR & SSIM & ELPIPS\\
\midrule
 Bi et al.
 &24.70 &0.958 &0.465  &20.92 &0.921 &0.783
 &27.29 &0.960 &0.378  &29.40 &0.971 &0.334
 &29.47 &0.970 &0.299\\
 NeRV
 &25.20 &0.960 &0.540  &25.68 &0.949 &0.526
 &27.67 &0.969 &0.306  &26.76 &0.968 &0.419
 &28.69 &0.969 &0.315\\
 Ours
 &{\bf 34.20} &{\bf 0.983} &{\bf 0.184}  &{\bf 33.51} &{\bf 0.974} &{\bf 0.302}  &{\bf 34.75} &{\bf 0.980} &{\bf 0.189}  &{\bf 33.86} &{\bf 0.981} &{\bf 0.257}  &{\bf 33.77} &{\bf 0.975} &{\bf 0.245}\\
\bottomrule
\end{tabular}
}
\end{center}
\vspace{-1em}
\end{table}

\begin{table}
\begin{center}
\small
\caption{\label{tab:env_quality} Quantitative comparisons on the test data for training on the ``env + point'' datasets. }
\vspace{-0.5em}
\setlength{\tabcolsep}{3pt}
\resizebox{\textwidth}{!}{
\begin{tabular}{l ccc ccc ccc ccc ccc} % lccc ccc ccc ccc ccc
\toprule
{\small{Env+Point}} & \multicolumn{3}{c}{Cow}  &  \multicolumn{3}{c}{Cloud} & \multicolumn{3}{c}{Bunny4-VaryA} & \multicolumn{3}{c}{Bunny4-VaryG} & \multicolumn{3}{c}{Buddha3}\\
\cmidrule(lr){2-4}\cmidrule(lr){5-7}\cmidrule(lr){8-10}\cmidrule(lr){11-13}\cmidrule(lr){14-16}
Method & PSNR & SSIM &ELPIPS & PSNR  & SSIM &ELPIPS &PSNR & SSIM & ELPIPS &PSNR & SSIM & ELPIPS&PSNR & SSIM & ELPIPS\\
\midrule
 Bi et al.
 &24.84 &0.960 &0.501  &22.18 &0.934 &0.709
 &26.65 &0.958 &0.464  &30.03 &0.974 &0.285
 &23.41 &0.938 &0.679\\
 NeRV
 &27.83 &0.974 &0.413  &26.07 &0.950 &0.476
 &28.18 &0.968 &0.301  &27.97 &0.975 &0.339
 &28.99 &0.969 &0.299\\
 Ours
 &{\bf 33.32} &{\bf 0.982} &{\bf 0.209}  &{\bf 32.64} &{\bf 0.969} &{\bf 0.353}  &{\bf 34.47} &{\bf 0.979} &{\bf 0.191}  &{\bf 34.09} &{\bf 0.982} &{\bf 0.243}  &{\bf 34.03} &{\bf 0.975} &{\bf 0.261}\\
\bottomrule
\end{tabular}
}
\end{center}
\vspace{-1em}
\end{table}

\paragraph{Relighting comparisons.} We show the qualitative comparisons of simultaneous relighting and view synthesis on the test data in Fig.~\ref{fig:relight_compare}. The left half is trained with the ``point'' lighting condition, whereas the right half is trained with the ``env + point''. 
Bi's method shows artifacts in each case as it has no mechanisms to simulate the multiple scattering that significantly affects the appearance of participating media. 
NeRV handles the environment illumination properly but shows artifacts on the participating media objects. Our method achieves realistic results on all test sets with either homogeneous media or heterogeneous media. 
Table~\ref{tab:point_quality} and Table~\ref{tab:env_quality} present the corresponding quantitative measurements, where our method overtakes the compared methods on each test set.

Using the same batch size, our training with 200K iterations on a Nvidia Quadro RTX 8000 GPU takes one day, whereas Bi's method and NeRV takes 22h and 46h. For a {\small{$400\times 400$}} image, our average inference time is $7.9$s, while Bi's method and NeRV takes $53.2$s and $21.9$s, respectively.

\paragraph{Learned lighting decomposition.} 
\begin{wrapfigure}{R}{0.7\textwidth}
\centering
\vspace{-1em}
\includegraphics[width=1.0\linewidth]{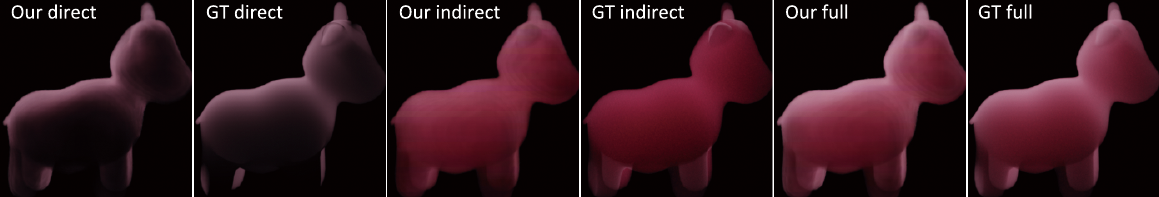}
\vspace{-1em}
\caption{\label{fig:light_decompose} Lighting decompositions.}
\vspace{-1em}
\end{wrapfigure}
Without using any ground-truth lighting decomposition data, our method is able to learn the decomposition of lighting in an unsupervised way. Figure~\ref{fig:light_decompose} presents our decomposed results on a test view of the \scene{Cow} scene,  with the single-scattering component (direct lighting) and the multiple-scattering component (indirect lighting), and the corresponding ground-truth images. 

\begin{wrapfigure}{R}{0.45\textwidth}
\centering
\vspace{-1em}
\includegraphics[width=1.0\linewidth]{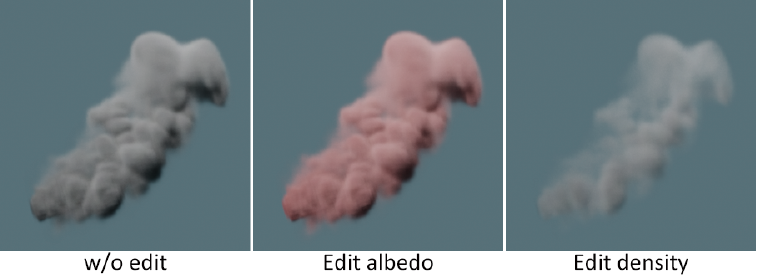}
\vspace{-1.25em}
\caption{\label{fig:scene_editing} Edit the learned cloud.}
\vspace{-1em}
\end{wrapfigure}

\paragraph{Scene editing and scene composition.} 
\begin{wrapfigure}{R}{0.45\textwidth}
\centering
\vspace{-1em}
\includegraphics[width=1.0\linewidth]{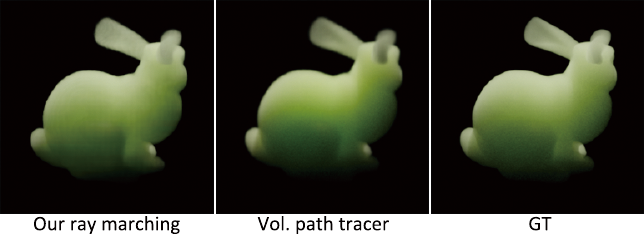}
\vspace{-1.5em}
\caption{\label{fig:compare_rm_vpt} Ray marching vs. volume path tracing.}
\vspace{-1em}
\end{wrapfigure}
Our method learns neural representations for the participating media scenes. After training, we can query the neural networks to obtain the volume density, the albedo, and the phase function parameter. This allows flexible editing to achieve desired effects or insertion into a new virtual environment for content creation. In addition, we can leverage a standard rendering engine to render these data. Figure~\ref{fig:compare_rm_vpt} compares the rendering of the learned \scene{Bunny} with ray marching using the neural network and with a volumetric path tracer. To render with the path tracer, we first queried the neural network to obtain 128 x 128 x 128 data volumes of volume density and albedo. Both rendered results are visually similar to the ground truth.
Figure~\ref{fig:scene_editing} demonstrates an editing of the red channel of the albedo to achieve the red cloud and another editing of the volume density to make the cloud thinner.

\begin{wrapfigure}{R}{0.45\textwidth}
\centering
\vspace{-1em}
\includegraphics[width=1.0\linewidth]{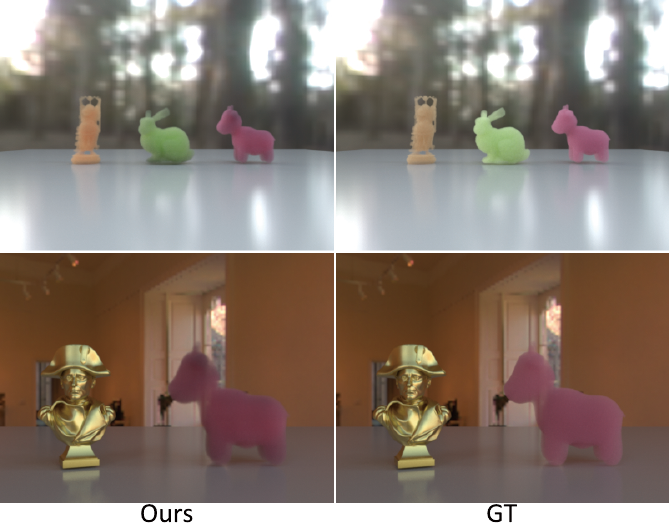}
\vspace{-1.5em}
\caption{\label{fig:scene_composition}Scene compositions.}
\vspace{-1.5em}
\end{wrapfigure}

We show in Fig.~\ref{fig:scene_composition} that we can compose a scene consisting of our learned cow and a gold sculpture described by traditional meshes and materials (Fig.~\ref{fig:scene_composition} bottom). Similarly, we can construct a scene composed entirely of our learned objects (Fig.~\ref{fig:scene_composition} top). To render the composed scenes, we slice out discrete volumes with a resolution $128\times 128\times 128$ from the volume density field and the albedo field and conduct the Monte Carlo rendering in Mitsuba~\cite{Mitsuba}.

\begin{figure}
\centering
\includegraphics[width=0.85\textwidth]{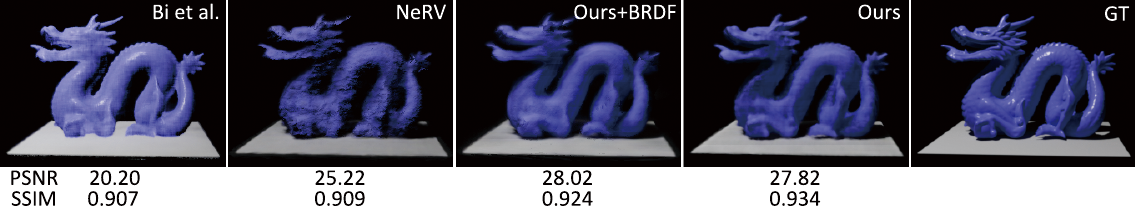}
\vspace{-1em}
\caption{\label{fig:dragon} Comparisons on a test view of the solid \scene{Dragon} with glossy surfaces. PSNR and SSIM metrics for this view are listed below images. The training illumination is from a single point light and the test illumination is a new white point light.}
\end{figure}

\paragraph{Scene of solid objects.} Beyond the scenes with participating media, our method can be used for scenes with solid objects. Figure~\ref{fig:dragon} shows a comparison between our method and the baselines on the \scene{Dragon} scene which contains glossy opaque surfaces. ``Ours+BRDF'' is a variant of our method that uses SH for indirect illumination, but adopts a classical BRDF model~\cite{karis2013real} and trains the neural network to predict parameters of the BRDF model as in~\cite{bi2020neural, srinivasan2020nerv}. Bi's method produces an overexposed appearance and the shadow on the floor gets faint. Our method achieves a smooth appearance and higher numerical metrics, whereas ``Ours + BRDF'' recovers the highlights on the glossy dragon better.

\subsection{Ablation Studies} \label{sec:ablation}
\begin{figure}
\centering
\begin{minipage}{0.45\textwidth}
\includegraphics[width=1.0\linewidth]{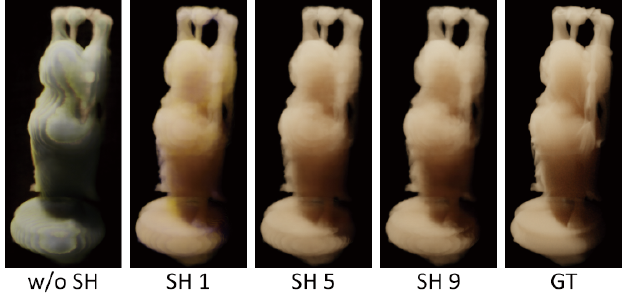}
\end{minipage}
\begin{minipage}{0.52\textwidth}
\small
\setlength{\tabcolsep}{6pt}
\renewcommand{\arraystretch}{0.9}
\begin{tabular}{lcccc}
\toprule
&PSNR$\uparrow$ &SSIM$\uparrow$ &ELPIPS$\downarrow$ &Time (s)\\
\midrule
w/o SH &25.91 &0.9301 &0.683 &3.91\\
SH-1 &32.53 &0.9728 &0.175 &5.16\\
SH-3 &32.70 &0.9739 &0.159 &5.75\\
SH-5 &{\bf 32.87} &{\bf 0.9743} & 0.154 &7.91\\
SH-7 &32.80 &0.9740 &{\bf 0.152} &13.00\\
SH-9 &32.61 &0.9739 &0.157 &16.58\\
\bottomrule
\end{tabular}
\end{minipage}
\caption{\label{fig:sh_ablation} Image quality and mean inference timings of different number of spherical harmonic bands. ELPIPS metrics have a scale of $10^{-2}$. The full images, including the SH-3 and SH-7, are documented in Appendix E.}
\vspace{-1em}
\end{figure}

\paragraph{Spherical harmonic bands.} We analyze the effect of the maximum spherical harmonics band $l_{max}$ of Eq.~\ref{eq:multiplescatter} based on the \scene{Buddha} scene. Figure~\ref{fig:sh_ablation} compares the same test view of each case on the left and tabulates the average quality measurements over the test set on the right. Removing the spherical harmonics from our method leads to quality drop and color shift is observed in its result. Based on the numerical metrics and inference timings, we select the $l_{max}=5$ for other experiments.

\paragraph{Scattering function.} 
\begin{wrapfigure}{R}{0.45\textwidth}
\centering
\vspace{-1em}
\includegraphics[width=1.0\linewidth]{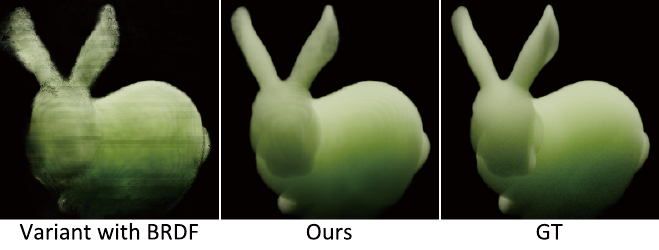}
\vspace{-1.5em}
\caption{\label{fig:Ablation_phase_brdf} Compare Ours+BRDF and Ours with phase function.}
\vspace{-1em}
\end{wrapfigure}
We show in Fig.~\ref{fig:dragon} that our method with the HG phase function can be applied to a scene of solid objects. In addition, we conduct an ablation by applying the variant ``Ours+BRDF'' to the \scene{Bunny} scene of participating media. Figure~\ref{fig:Ablation_phase_brdf} shows that the variant has difficulty in learning the volume density and leads to many cracks, while the proposed method performs robustly on this scene.

\section{Limitations and Future Work} \label{sec:limitations}

\paragraph{Real-world scenes.} In this work, we learn neural representations from synthetic datasets with varying but known lighting conditions. Also, the camera poses are available to the method. It would be interesting to extend this method to handle participating media captured from real-world scenes with unknown lighting conditions and unknown camera poses. In that case, the illumination and camera poses of the scenes need to be estimated in the first place.

\paragraph{Glossy reflections.} For scenes with glossy solid objects (Fig.~\ref{fig:dragon}), our method tends to reproduce a smooth appearance and the glossy highlights are not as sharp as the ground truth. An avenue for future research would be to develop methods to recover the glossy reflections better.

\paragraph{Generalizability.} Our ray marching with the trained neural network generalizes well to unseen light intensities and light locations that are in the range of the training data. That said, its generalization is in an interpolation manner. For light intensities and light locations that are outside of the training range, the generalization quality of the neural network gradually decreases. Please refer to the Appendix H for the analysis on the generalization quality.

\paragraph{Media within refractive boundaries.} Our method achieves realistic results for participating media without refractive boundaries, like cloud, fog, and wax figure. Applying our method to participating media within refractive boundaries, like wine in a glass, entails further work as the refractive boundaries cause ambiguities due to deflecting the camera rays and the light rays.
\section{Conclusion}

We have proposed a novel method for participating media reconstruction from observed images with varying but known illumination. We propose to simulate direct illumination with Monte Carlo ray tracing and approximate indirect illumination with learned spherical harmonics. This enables our approach to learn to decompose the illumination as direct and indirect components in an unsupervised manner. Our method learns a disentangled neural representation with volume density, scattering albedo and phase function parameters for participating media, and we demonstrated its flexible applications in relighting, scene editing and scene compositions.

\begin{ack}
We acknowledge the valuable feedback from reviewers. This work was supported by Research Executive
Agency 739578 and CYENS Phase 2 AE739578.
\end{ack}

%%%%%%%%%%%%%%%%%%%%%%%%%%%%%%%%%%%%%%%%%%%%%%%%%%%%%%%%%%%%
\section*{Appendices}
\appendix
\section{Phase Function Details}

Our phase function for participating media (Sec.~\ref{sec:lightdecompose}) is the Henyey-Greenstein (HG) function~\cite{henyey1941diffuse}
\begin{equation}\label{eq:phase}
    p(\omega_{o}, \omega_{i}, g) = \frac{1}{4\pi}\frac{1-g^{2}}{\left(1+g^{2}+2g \cos{\theta}\right)^{3/2}},
\end{equation}
where $\theta$ is the angle between the outgoing direction $\omega_{o}$ and the incident direction $\omega_{i}$. For the notations of directions, we use the convention that both the incident and outgoing rays point away from a scattering location. $g$ is called the \textit{asymmetry parameter} that is in $(-1, 1)$. A positive $g$ value is for forward scattering, a negative $g$ value is for backward scattering, and the zero value is for isotropic scattering. 

Same as in~\cite{bi2020neural, srinivasan2020nerv}, our BRDF function for the experiment on solid objects (Sec.~\ref{sec:results}) is the analytical model~\cite{karis2013real} which combines a specular component using the ggx distribution~\cite{walter2007microfacet} and a diffuse component.

\section{Spherical Harmonics Details}

In Sec.~\ref{sec:lightdecompose} of the main paper, we propose to represent the incident radiance due to multiple scattering  with spherical harmonics. Spherical Harmonics (SH) are orthonormal basis defined on complex numbers over the unit sphere. Since our radiance function is defined in the real number domain, our SH basis functions $Y_l^{m}(\omega_i) \left(0\leq l\leq l_{max}, -l\leq m\leq l\right)$ in Eq.~\ref{eq:multiplescatter} are \textit{real} spherical harmonic functions
\begin{equation}
  Y_{l}^{m}(\theta_i, \phi_i) =
    \begin{cases}
      \sqrt{2}K_{l}^{m}\cos(m\phi_i)P_{l}^{m}(\cos\theta_i) & \text{$m > 0$}\\
      K_{l}^{m}P_{l}^{m}(\cos\theta_i) & \text{$m = 0$}\\
      \sqrt{2}K_{l}^{m}\sin(-m\phi_i)P_{l}^{-m}(\cos\theta_i) & \text{$m < 0$,}
    \end{cases}       
\end{equation}
where $(\theta_i, \phi_i)$ are the spherical coordinates of the direction $\omega_i$ that is in the Cartesian coordinate system, $K_{l}^{m} = \sqrt{\frac{(1+2l)}{4\pi}\frac{(l-\left|m\right|)!}{(l+\left|m\right|)!}}$ is a normalization factor, and $P_{l}^{m}$ are the associated Legendre polynomials. Then, the incident radiance function $\widetilde{L}(\omega_i)$ can be computed using the SH basis
\begin{equation}
\widetilde{L}(\omega_i) = \mathcal{F}\left(\sum_{l=0}^{l_{max}}\sum_{m=-l}^{l}c_{l}^{m}Y_{l}^{m}(\omega_i)\right),
\end{equation}
where $c_{l}^{m}\in\mathbb{R}^3$ are SH coefficients, $l_{max}$ is the maximum SH band, and $\mathcal{F}(x)=\max(0, x)$ ensures non-negative incident radiance. 

In each training iteration, we sample $K=64$ random incident directions $\{\omega_i\}_{i=1}^{K}$ and evaluate the $Y_{l}^{m}(\omega_i)$. To reduce the computational cost, we reuse these incident directions and the evaluated $Y_{l}^{m}(\omega_i)$ for all point samples along the primary rays of this batch.

\section{Additional Implementation Details of Our Method}

\paragraph{Training details.} 
We end-to-end train our model to learn a separate neural representation of each scene. In each training iteration, we randomly draw a batch of $1200$ primary rays across all training views. Our visibility network is trained to match the learned scene geometry by the property network, so it is optimized according to the visibility values computed from the volume density, without requiring ground-truth visibility. We cut off the gradient from the \textit{render loss} to the visibility network. Meanwhile, we cut off the gradient from the \textit{visibility loss} to the property network so that it does not degrade its learning of the volume density.

\paragraph{Inference details.} Our inference uses the same setting as the training. We draw $64$ point samples along each camera ray to query our model. The number of incident directions for computing indirect illumination is $64$. We set the number of shadow rays to $1$ for the point light, whereas we use $32$ shadow rays for the environment lighting.

\section{Implementation details of the two baselines}\label{sec:appendix_detail}

For our comparisons, we implemented the Neural Reflectance Field~\cite{bi2020neural} and NeRV~\cite{srinivasan2020nerv} as our baselines. Since they were designed for scenes with solid objects, we adapt them to cope with participating media.

Our implementation of the Neural Reflectance Field~\cite{bi2020neural} baseline uses the same neural network architecture and positional encoding as in the original paper. Specifically, we implement the dual-network design with a coarse network and a fine network. Each neural network is an MLP consisting of $14$ fully-connected ReLU layers with $256$ neurons per layer. Also, we apply frequency-based positional encoding to transform the input 3D coordinates with a maximum frequency $2^{10}$. Different from the eight-channel output in the original paper, we have a four-channel output consisting of a 1-D volume density and a 3-D scattering albedo. Along each ray, we draw $64$ stratifed point samples for the coarse network and $128$ point samples for the fine network.

In our implementation of the NeRV~\cite{srinivasan2020nerv} baseline, we utilize an MLP with $8$ fully-connected ReLU layers to compute the physical properties. Each layer has $256$ neurons. In addition, we employ a visibility MLP~\cite{srinivasan2020nerv} to compute a 1-D visibility and a 1-D expected termination depth. The visibiliy MLP firstly processes the encoded coordinates using $8$ fully-connected ReLU layers with $256$ neurons per layer to get an 8-D output. The output, concatenated with the encoded directions, is further processed by $4$ fully-connected ReLU layers with $128$ neurons per layer. The maximum positional encoding frequencies for 3D coordinates and directions are $2^{7}$ and $2^{4}$, respectively. Along each camera ray, we take $64$ stratified samples, same as in our method. We trace one shadow ray for the point light source and $32$ shadow rays for the environment lighting. For the ``point'' illumination, we uniformly take $128$ random directions for the first indirect bounces at the termination depth along a ray. For the ``env + point'' illumination, we set the number of first indirect bounces to $32$.  

\section{Additional images for the spherical harmonic band ablation}

\begin{figure}[htbp]%[width=0.85\textwidth]\textbf{}
\centering
\includegraphics{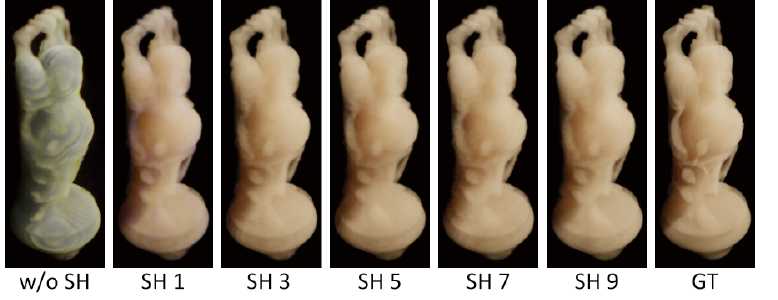}
%\vspace{-1em}
\caption{\label{fig:SH_ablation_full} Qualitative comparisons of different settings of the maximum spherical harmonic band from SH-$1$ to SH-$9$. ``w/o SH'' denotes without using spherical harmonics and ``GT'' denotes the ground truth.}
\end{figure}

In the ablation study of the maximum spherical harmonic band (Sec.~\ref{sec:ablation}), we show the results of SH-$1$, SH-$5$, and SH-$9$. We present the full comparison with SH-3 and SH-5 in Fig.~\ref{fig:SH_ablation_full}. The quantitative metrics are listed in the right table of Fig.~\ref{fig:sh_ablation} (Sec.~\ref{sec:ablation}). Note that SH-5 achieves the  qualitative result that is similar to SH-7 and SH-9, but SH-5 gives higher numerical performance than others (Fig.~\ref{fig:SH_ablation_full}).

\section{Additional quantitative results for participating media scenes}

\subsection{Scenes trained on the ``point''}
Table~\ref{tab:quantitative_point} presents additional numerical results for the single \scene{Bunny} and \scene{Buddha} scenes when using the ``point'' training illumination. Each test view has a new point light. The tabulated values are the mean values over images of the test set. Our approach outperforms the two baselines quantitatively.

\begin{table}
\begin{center}
\small
\caption{\label{tab:quantitative_point} Quantitative comparisons on the test data. Image qualities are measured with PSNR ($\uparrow$), SSIM ($\uparrow$) and ELPIPS ($\downarrow$)~\cite{kettunen2019lpips}. ELPIPS values have a scale of $\times 10^{-2}$. }
%\vspace{-1em}
%\setlength{\tabcolsep}{3pt}
\renewcommand{\arraystretch}{1.0}
%\resizebox{\textwidth}{!}{
\begin{tabular}{l ccc ccc} % lccc ccc ccc ccc ccc
\toprule
{\small{Point}}& \multicolumn{3}{c}{Bunny} &  \multicolumn{3}{c}{Buddha}\\
\cmidrule(lr){2-4}\cmidrule(lr){5-7}
Method & PSNR & SSIM &ELPIPS & PSNR  & SSIM &ELPIPS\\
\midrule
Bi et al.  
&22.46 &0.933 &0.720 &23.62 &0.951 &0.518
\\
NeRV   
&24.57 &0.951 &0.627 &25.47 &0.959 &0.432
\\
Ours
&{\bf 33.49} &{\bf 0.982} &{\bf 0.209} &{\bf 32.87} &{\bf 0.974} &{\bf 0.154}
\\
\bottomrule
\end{tabular}
%}
\end{center}
% \vspace{-1em}
\end{table}

\subsection{Scenes trained on the ``env + point''} 

Accordingly, Table~\ref{tab:quantitative_env} presents quantitative results for the \scene{Bunny} and the \scene{Buddha} scenes when they are trained with the ``env + point'' illumination. Each test view has a single novel point light. Our method numerically performs better than the baselines. 
\begin{table}[h]
\begin{center}
\small
\caption{\label{tab:quantitative_env} Quantitative comparisons on the \scene{Bunny} and the \scene{Buddha} scene when they are trained with the ``env + point'' illumination. The PSNR ($\uparrow$), SSIM ($\uparrow$) and ELPIPS ($\downarrow$)~\cite{kettunen2019lpips} values are averaged over all images of a test set. ELPIPS values have a scale of $\times 10^{-2}$. }
%\vspace{-1em}
%\setlength{\tabcolsep}{3pt}
\renewcommand{\arraystretch}{1.0}
%\resizebox{\textwidth}{!}{
\begin{tabular}{l ccc ccc} % lccc ccc ccc ccc ccc
\toprule
{\small{Env+Point}}& \multicolumn{3}{c}{Bunny} & \multicolumn{3}{c}{Buddha}\\
\cmidrule(lr){2-4}\cmidrule(lr){5-7}
Method & PSNR & SSIM &ELPIPS & PSNR  & SSIM &ELPIPS \\
\midrule
Bi et al.  
&22.82 &0.935 &0.709 &24.20 &0.956 &0.466
\\
NeRV   
&25.24 &0.959 &0.597 &27.36 &0.968 &0.345
\\
Ours
&{\bf 32.93} &{\bf 0.980} &{\bf 0.293} &{\bf 32.74} &{\bf 0.976} &{\bf 0.204}
\\
\bottomrule
\end{tabular}
%}
\end{center}
% \vspace{-1em}
\end{table}

\section{Quantitative results for scenes of solid objects}

In Table~\ref{tab:solid}, we show the quantitative results on the \scene{Dragon} scene and the \scene{Armadillo} scene that contain glossy solid objects. Bi's method and the NeRV method use the parameter settings as described in Appendix~\ref{sec:appendix_detail}. Our method retains the same parameter settings as those for the participating media scenes. ``Ours + BRDF'' is a variant of our method that uses a BRDF function as the scattering function. We experimentally set its maximum spherical harmonic band to $1$ and set the highest positional encoding frequency to $2^{7}$. Our method achieves higher quantitative metrics compared to the baselines and the variant. 

Figure~\ref{fig:dragon} of the main paper shows a comparison on one test view of the \scene{Dragon} scene. Bi's method recovers the highlights on surfaces, but it produces an overexposed appearance and leads to faint shadows. Our method produces a smooth appearance and properly cast the shadow according to the novel lighting.
\begin{table}[tbp]
\begin{center}
\small
\caption{\label{tab:solid} Quantitative comparisons on two scenes with glossy solid objects. The scenes are trained with the ``point'' illumination and tested under novel lighting. The PSNR ($\uparrow$), SSIM ($\uparrow$) and ELPIPS ($\downarrow$)~\cite{kettunen2019lpips} are the averaged value over a test set. ELPIPS values have a scale of $\times 10^{-2}$. }
%\vspace{-1em}
%\setlength{\tabcolsep}{3pt}
\renewcommand{\arraystretch}{1.0}
%\resizebox{\textwidth}{!}{
\begin{tabular}{l ccc ccc} % lccc ccc ccc ccc ccc
\toprule
& \multicolumn{3}{c}{Dragon} & \multicolumn{3}{c}{Armadillo}\\
\cmidrule(lr){2-4}\cmidrule(lr){5-7}
Method & PSNR & SSIM &ELPIPS & PSNR  & SSIM &ELPIPS \\
\midrule
Bi et al.
&19.60 &0.897 &1.358  & 19.19 &0.897 &1.243
\\
NeRV   
&26.60 &0.917 &0.902  & 25.32 &0.893 &1.034
\\
Ours + BRDF 
&28.32 &0.931 &0.692  & 26.46 &0.917 &0.847
\\
Ours
&{\bf 28.50} &{\bf 0.942} & {\bf 0.564} &{\bf 26.60} &{\bf 0.924} &{\bf 0.753}
\\
\bottomrule
\end{tabular}
%}
\end{center}
% \vspace{-1em}
\end{table}

\section{Generalization quality}
\begin{figure}
\centering
\includegraphics[width=1.0\textwidth]{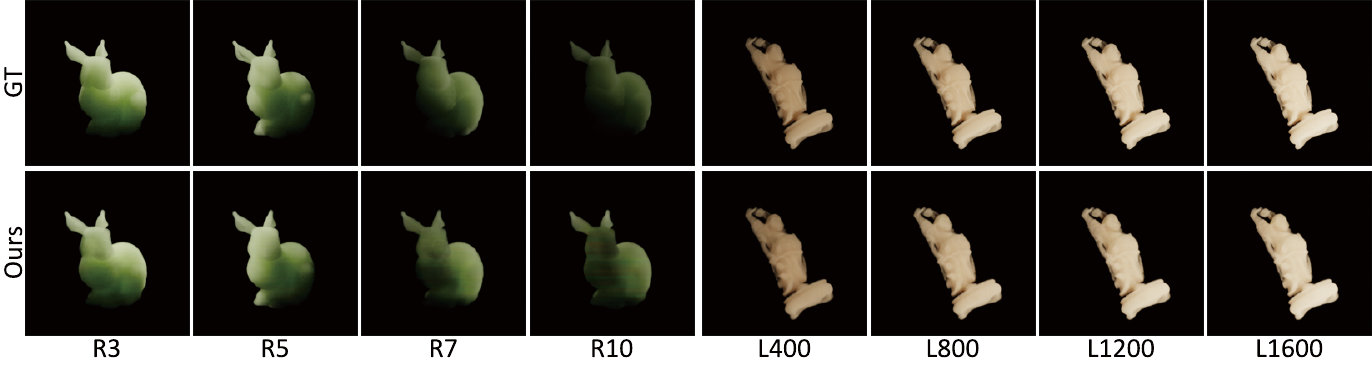}
\vspace{-1.5em}
\caption{\label{fig:generalization} Qualitative comparisons of generalization on light location and light intensity.}
\end{figure}

\begin{figure}[th!]
\centering
\begin{minipage}{0.49\textwidth}
\small
\setlength{\tabcolsep}{6pt}
\renewcommand{\arraystretch}{0.9}
\begin{tabular}{l cccc}
\toprule
Bunny &\multicolumn{4}{c}{Light distance (m)}\\
\cmidrule(lr){2-5}
&3 &5 &7 &10\\
\midrule
PSNR &31.39 &29.87 &23.14 &16.96\\
SSIM &0.970 &0.965 &0.922 &0.865\\
ELPIPS &0.359 &0.388 &0.923	&1.740\\
\bottomrule
\end{tabular}
\end{minipage}
\begin{minipage}{0.49\textwidth}
\small
\setlength{\tabcolsep}{6pt}
\renewcommand{\arraystretch}{0.9}
\begin{tabular}{l cccc}
\toprule
Buddha &\multicolumn{4}{c}{Light intensity}\\
\cmidrule(lr){2-5}
&400 &800 &1200 &1600\\
\midrule
PSNR &32.82 &33.04 &31.91 &28.47\\
SSIM &0.982	&0.984 &0.981 &0.974\\
ELPIPS &0.204 &0.195 &0.219 &0.235\\
\bottomrule
\end{tabular}
\end{minipage}
\caption{\label{fig:generalization_quantitative} Quantitative results for investigating the generalization on the light locations and light intensities. ELPIPS metrics have a scale of $10^{-2}$.}
\vspace{-1em}
\end{figure}
\paragraph{Generalization quality for light distance.} We used the \scene{Bunny} scene in this experiment. During training, the point light's distance to the center of the bunny is stratified sampled from the range $(3, 5)$. We then build four test sets, each with 20 views; we set the point light's distance for the test sets as $3$, $5$, $7$, and $10$, individually. The light intensity is set to 600. We run the trained neural network on each test set. The image results are presented on the left of Fig.~\ref{fig:generalization} and the numerical performance is shown in the left table of Fig.~\ref{fig:generalization_quantitative}.
The trained neural network performs relatively well when the test point light's distance is close to those in training, and the numerical performance gradually drops when the test point light moves away from the training manifold.

\paragraph{Generalization quality for light intensity.} We used the \scene{Buddha} scene in this experiment. The training intensity values were stratified sampled from $50$ to $900$. We generate four test sets with the same 20 camera views. The point light is put at a distance $4$ for all test sets. Also, we set the test light intensity as $400$, $800$, $1200$, and $1600$, respectively. Same as before, we run the trained neural network on each test set. The graphical comparisons and numerical metrics are shown on the right of Fig.~\ref{fig:generalization} and Fig.~\ref{fig:generalization_quantitative}.
We observed that the trained neural network achieves high numerical performance when the test light intensity is within the range of the training intensity. For testing intensity outside of the training range, the numerical performance decreases.

% {\small
% \bibliographystyle{unsrtnat}
% \bibliography{neurips}
% }

%%%%%%%%%%%%%%%%%%%%References%%%%%%%%%%%%%%%%%%%%%%%%%%
{
\small
\bibliographystyle{unsrtnat}
\bibliography{main}
}
% \section*{References}
% References follow the acknowledgments. Use unnumbered first-level heading for
% the references. Any choice of citation style is acceptable as long as you are
% consistent. It is permissible to reduce the font size to \verb+small+ (9 point)
% when listing the references.
% Note that the Reference section does not count towards the page limit.
% \medskip

% {
% \small
% [1] Alexander, J.A.\ \& Mozer, M.C.\ (1995) Template-based algorithms for
% connectionist rule extraction. In G.\ Tesauro, D.S.\ Touretzky and T.K.\ Leen
% (eds.), {\it Advances in Neural Information Processing Systems 7},
% pp.\ 609--616. Cambridge, MA: MIT Press.

% [2] Bower, J.M.\ \& Beeman, D.\ (1995) {\it The Book of GENESIS: Exploring
%   Realistic Neural Models with the GEneral NEural SImulation System.}  New York:
% TELOS/Springer--Verlag.

% [3] Hasselmo, M.E., Schnell, E.\ \& Barkai, E.\ (1995) Dynamics of learning and
% recall at excitatory recurrent synapses and cholinergic modulation in rat
% hippocampal region CA3. {\it Journal of Neuroscience} {\bf 15}(7):5249-5262.
% }

\end{document}